# Stemmers for Tamil Language: Performance Analysis

M.Thangarasu
M.Phil Scholar, Full Time,
Department of Computer Science and Application
K.S.Rangasamy College of Arts and Science
Tiruchengode, India
E-mail:thangarasumathan@gmail.com

Dr.R.Manavalan
Department of Computer Science and Application
K.S.Rangasamy College of Arts and Science
Tiruchengode, India
E-mail: manavalan_r@rediffmail.com

**Abstract— Stemming is the process of extracting root word from the given inflection word and also plays significant role in numerous application of Natural Language Processing (NLP). Tamil Language raises several challenges to NLP, since it has rich morphological patterns than other languages. The rule based approach light-stemmer is proposed in this paper, to find stem word for given inflection Tamil word. The performance of proposed approach is compared to a rule based suffix removal stemmer based on correctly and incorrectly predicted. The experimental result clearly show that the proposed approach light stemmer for Tamil language perform better than suffix removal stemmer and also more effective in Information Retrieval System (IRS).**

## I. INTRODUCTION

Tamil is a Dravidian language. It is the regional language of Tamil Nadu of India. It has number of morphological variant for a word. For example ÀÊò¾¡ý (a male who has read), ÀÊì¸¢È¡ý(a male who is reading now), ÀÊôÀ¡ý (a male who will read), ÀÊì¸Á¡ð¼¡ý(a male who will not read). This creates more complexity in information retrieval. Stemmer is especially used in Information Retrieval System (IRS) for improving their performance at the mean time reduce the complexity, for example when a user enter query word search (§¾Î), user most likely wants to retrieve documents containing the terms searching (§¾Îõ) and searched (§¾ÊÂ) etc. as well. Thus using stemmer improves recall (i.e.) the number of documents retrieved in response to a query, since many terms are mapped to one. The benefit of Stemmer is decreases the size of the index files in the IR system. Consequently many stemmers algorithm have been proposed and evaluated for various Indian languages such as Hindi [1], Gujarati [2], and Punjabi [3] etc. An overview of the proposed model projected in Figure 1.

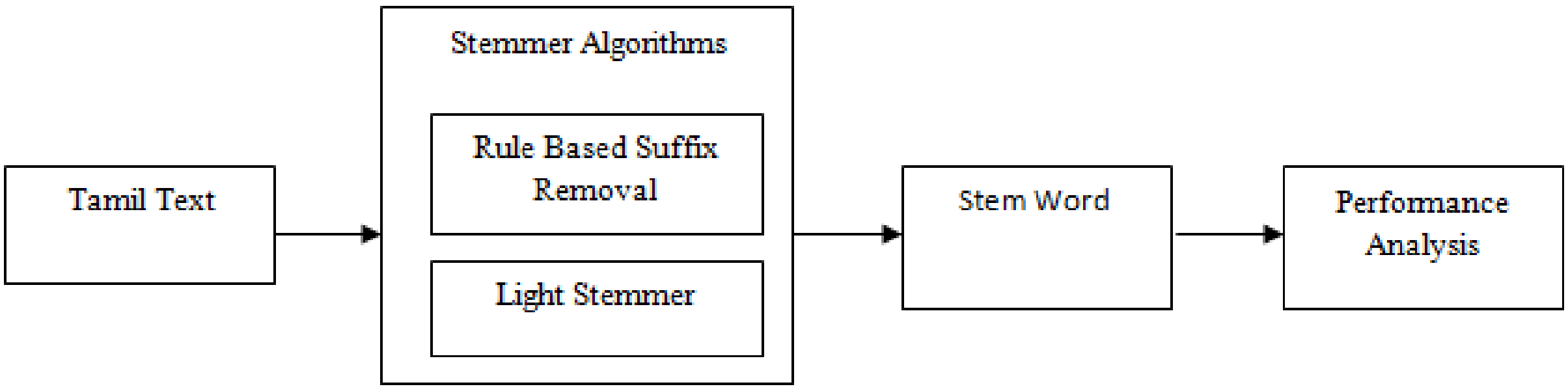

Figure1. Overview of Proposed Model for Tamil Stemmer Algorithms

The proposed light stemmer algorithm compared with rule based suffix removal algorithm. This paper organized as follows: Section II describes the related work. Section III discusses about the Tamil language Section IV presents the proposed system light stemmer and also describes a rule based suffix removal stemmer. Section V and VI contains an experimental analysis and conclusion of the work respectively.

## II. RELATED WORK

Earlier, Stemmer was primarily developed for English, but later due to the corpus growth of languages other than English, there was an increased demand from the research community to develop stemmers for other languages too. In case in Indian languages, the earliest work reported by Ramanathan and Rao [1] (2003) to perform longest match stripping for building a Hindi stemmer. Juhi Ameta et al. developed a light stemmer for Gujarati language [2] in 2011 for removing inflectional and derivational endings in order to reduce word forms to common stem. Slowly investigation starts for other language such as Bengali [4], Urdu [5]**,** Malayalam [6] and Punjabi [3] are also carried out. However, in this section express research experience in developing Tamil stemmers.

The most common approaches used for developing stemmers are Brute force and Affix strip. Among all the existing approaches, this research tries to implement a light stemmer for Tamil for inherent to develop a stemmer algorithm in an easier and faster way.

## III. TAMIL LANGUAGE

Tamil is a Dravidian language spoken predominantly by Tamil people of Indian subcontinent. Tamil words typically have more morphological patterns than English words. Tamil word contains constituent parts: a stem, which can be thought of as responsible for the nuclear meaning of verb, attached to which may be zero or more derivational prefix and zero or one suffix, which together to form a word. Tamil is a morphologically rich language resulting in its relatively high inflectional forms. Normally most of the Tamil words have more than one morphologically suffixes. The number of suffix is ranging from 3 to 13. The description with example nouns, compound nouns and verbs are given in the following sequel.

### *A. Noun*

Tamil has an extensive case system. Root nouns can assume eight different morphological shapes depending on their roles in a sentence. Singular and plural forms are also distinguished through inflections. Suffixes are attached to stem word of the noun. Table I shows example for different stem noun.

TABLE I NOUN

| | | |
|---|---|---|
| Singular | PeN(¦Àñ)/ girl | maram(ÁÃõ) /tree |
| Oblique stem | PeN(¦Àñ)- | maram(ÁÃõ) - |
| Nominative stem | PeN(¦Àñ) | maram(ÁÃõ) |
| Accusative stem | PeN(¦Àñ)-ai(³) | maram(ÁÃõ) -ai |
| Dative stem | PeN(¦Àñ)-ukku(¯ìÌ) | maram(ÁÃõ) -ukku(¯ìÌ) |
| Sociative stem | PeN(¦Àñ)-odu(µÎ) | maram(ÁÃõ) -odu(µÎ) |
| Genitive stem | PeN(¦Àñ)-udaiya(¯¨¼Â) | maram(ÁÃõ) -udaiya(¯¨¼Â) |
| Instrumental stem | PeN(¦Àñ)–aal(¬ø) | maram(ÁÃõ) -aal(¬ø) |
| Locative stem | PeN(¦Àñ)-idam(þ¼õ) | maram(ÁÃõ) -marath(þø) |
| Ablative stem | PeN(¦Àñ)–idamirunthu(þ¼Á¢ÕóÐ) | maram(ÁÃõ) -ilirunthu(þÄ¢ÕóÐ) |
| Vocative stem | PeN(¦Àñ)-e(±) | maram(ÁÃõ) -e(±) |
| Plural | PeNgaL(¦Àñ¸û)/ girls | marangal(ÁÃí¸û)/ trees |
| Oblique stem | PeNgaL(¦Àñ¸û)- | marangal(ÁÃí¸û)- |
| Nominative stem | PeNgaL(¦Àñ¸û) | marangal(ÁÃí¸û) |
| Accusative stem | PeNgaL(¦Àñ¸û)-ai(³) | marangal(ÁÃí¸û)-ai(³) |
| Dative stem | PeNgaL(¦Àñ¸û)-ukku(¯ìÌ) | marangal(ÁÃí¸û)-ukku(¯ìÌ) |
| Sociative stem | PeNgaL(¦Àñ¸û)-odu(µÎ) | marangal(ÁÃí¸û)-odu(µÎ) |
| Genitive stem | PeNgaL(¦Àñ¸û)-udaiya(¯¨¼Â) | marangal(ÁÃí¸û)-udaiya(¯¨¼Â) |
| Instrumental stem | PeNgaL(¦Àñ¸û)-aal(¬ø) | marangal(ÁÃí¸û)-aal(¬ø) |
| Locative stem | PeNgaL(¦Àñ¸û)-idam(þ¼õ) | marangal(ÁÃí¸û)-il(þø) |
| Ablative stem | PeNgaL(¦Àñ¸û)- idamirunthu(þ¼Á¢ÕóÐ) | marangal(ÁÃí¸û)-ilirunthu(þÄ¢ÕóÐ) |
| Vocative stem | PeNgaL(¦Àñ¸û)-e(±) | marangal(ÁÃí¸û)-e(±) |

* 'gaL'(¸û) is added to the singular form of nouns to make them plural

### *B. Compound Nouns*

Nouns are also occurring in various compound forms as well. It can be made up of several units where each unit expresses a particular grammatical meaning. The Tamil noun, “pAdikkoNdurunhthavanai (À¡Êì¦¸¡ñÎÕó¾Å¨É)”, which translates as, “the male who was singing”, gives information on tense, number, gender, person and case. This noun is actually derived from the full non-infinite verb, “pAdikkoNdu (À¡Êì¦¸¡ñÎ)”, which means, “singing”. In English, deriving nouns from verbs is seen too. The full finite verb, “sing”, for instance could be changed into a noun by adding the suffix “er” to its stem, so that it becomes “singer”. But, while Tamil is an agglutinating language, English is not.

### *C. Verbs*

Tamil verbs may be main or auxiliary. They also exist in finite and non-finite forms just as in English. Tamil finite verbs however give much more grammatical information than English finite verbs do, in that they mark mood, tense, number, person, gender, case etc… In the Table II below, can observe the different finite morphological construction for the verb “padi(ÀÊ)” [study].

TABLE II VERBS

| | **Past** | **Present** | **Future** | **Future-Neg** |
|---|---|---|---|---|
| I singular | padi(ÀÊ)- ththEn(ò§¾ý) | Padi(ÀÊ)- kkiREn(ì¸¢§Èý) | Padi(ÀÊ)- ppEn(ô§Àý) | Padi(ÀÊ)- kkamaattEn(ì¸Á¡ð§¼ý) |
| II singular | Padi(ÀÊ)- ththAi (ò¾¡ö) | Padi(ÀÊ)- kkiRaai(ì¸¢È¡ö) | Padi(ÀÊ)- ppaai(ôÀ¡ö) | Padi(ÀÊ)- kkamataai(ì¸Á¼¡ö) |
| III singular male | Padi(ÀÊ)- ththaan(ò¾¡ý) | Padi(ÀÊ)- kkiRaan(ì¸¢È¡ý) | Padi(ÀÊ)- ppaan(ôÀ¡ý) | Padi(ÀÊ)- kkamaattaan(ì¸Á¡ð¼¡ý) |
| III singular female | Padi(ÀÊ)-ththaaL (ò¾¡û) | Padi(ÀÊ)- kkiRaaL(ì¸¢È¡û) | Padi(ÀÊ)- papal(ôÀ¡û) | Padi(ÀÊ)- kkamaattaaL(ì¸Á¡ð¼¡û) |
| III singular hon | Padi(ÀÊ)- ththaar (ò¾¡÷) | Padi(ÀÊ)- kkiRaar(ì¸¢È¡÷) | Padi(ÀÊ)- ppaar(ôÀ¡÷) | Padi(ÀÊ)- kkamaatdaar(ì¸Á¡ð¼¡÷) |
| III singular inan | Padi(ÀÊ)- ththathu(ò¾Ð) | Padi(ÀÊ)- kkiRathu(ì¸¢ÈÐ) | Padi(ÀÊ)- kkum(ìÌõ) | Padi(ÀÊ)- kkaathu(ì¸¡Ð) |
| I plural | Padi(ÀÊ)- tthOm(ð§¾¡õ) | Padi(ÀÊ)- kkiROm(ì¸¢§È¡õ) | Padi(ÀÊ)- pPOm(ô§À¡õ) | Padi(ÀÊ)- kkamaatTOm ì¸Á¡ð§¼¡õ) |
| II plural | Padi(ÀÊ)-ththIrkaL (ò¾£÷¸û) | Padi(ÀÊ)- kkiRIrkaL(ì¸¢È£÷¸û) | Padi(ÀÊ)- ppIRkaL(ôÀ£ü¸û) | Padi-(ÀÊ)- kkamaattaarkaL(ì¸Á¡ðË÷¸û) |
| III plural an | Padi(ÀÊ)- ththaarkaL (ò¾¡÷¸û) | Padi(ÀÊ)- kkiRaarkaL(ì¸¢È¡÷¸û) | Padi(ÀÊ)- ppaarKaL(ôÀ¡÷¸û) | Padi(ÀÊ)- kkamaattaarkaL(ì¸Á¡ð¼¡÷¸û) |
| III plural inan | Padi(ÀÊ)- ththana(ò¾É) | Padi(ÀÊ)- kkinRana(ì¸¢ýÈÉ) | Padi(ÀÊ)- kkum(ìÌõ) | Padi(ÀÊ)- kkaathu(ì¸¡Ð) |

non-future negative: padi(ÀÊ)-kkavillai(ì¸Å¢ø¨Ä) (all persons, numbers and genders)*

TABLE III NON-FINITE VERB

| | |
|---|---|
| Conjunctive | Padi(ÀÊ)-thu (Ð) |
| Infinitive | Padi(ÀÊ)-kka( ì¸ ) |
| Neg. verbal participle | Padi(ÀÊ)-kkaamal( ì¸ ¡Áø) |
| Conditional | Padi(ÀÊ)-thaal(¾¡ø) |
| Neg. conditional | Padi(ÀÊ)-kkaanittaal( ì¸ É¢ð¼ø) |
| Neg.relative participle | Padi(ÀÊ)-kkaatha( ì¸ ¡¾) |
| Neg. verbal noun | Padi(ÀÊ)-kkaathathu<br>( ì¸ ¡¾Ð) |
| Deverbal nouns | Padi-(ÀÊ)thal( ¾¡ø);padi(ÀÊ)-ppu( ôÒ); padi(ÀÊ)-kkai( ì¨¸ ) |

## IV. STEMMERS FOR TAMIL LANGUAGE

Stemming is the process of extracting root word from the given text. Without performing complete morphological analysis, it is also reduces the total number of distinct index entries. Richness of morphology in Tamil language is major issue for creating right stemmer. To address of this issue, light stemmer is proposed and performance compared to rule based suffix stripping stemmer. The detailed description of two stemmer approaches for Tamil language such as rule based suffix removal stemmer and light stemmer are discussed following sequel.

### *A. Rule based suffix stripping stemmer algorithm*

Rule based suffix stripping stemmer algorithm truncate suffix of Tamil inflectional word based on the rules. That convert inflectional Tamil word into stemmed Tamil word [3][6].Rule based suffix stripping stemmer Algorithm for Tamil Language is presented in Figure 2.

Rule based suffix stripping stemmer Algorithm for Tamil Language

Input : List of Tamil words

Output : Stemmed(Root) words

Step1 : Eliminate the entire complex plural. Eg («Å÷¸û, ¦ºø¸¢È¡÷¸û, Åó¾¡÷¸û...)

«Å÷¸û= «Å÷, ¦ºø¸¢È¡÷¸û= ¦ºø¸¢È¡÷, Åó¾¡÷¸û= Åó¾¡÷

Step2 : Eliminate the join word suffixes. Eg («Å÷, ¦ºø¸¢È¡÷, Åó¾¡÷)

«Å÷= «Å÷, ¦ºø¸¢È¡÷= ¦ºø¸¢, Åó¾¡÷= Åó¾

A : Eliminate È¡÷

¦ºø¸¢È¡÷= ¦ºø¸¢

B : Eliminate ¡÷

Åó¾¡÷= Åó¾

Step 3 : According to the identified suffix, the next possible suffix list is generated using rules.

Figure 2. A Rule Based Suffix Stripping Algorithm for Tamil Language

### *B. Light Stemmer*

The rule based suffix stripping stemmer algorithm give infinite verb for some Tamil words. Light stemmer is proposed to address above issue. Light stemmer is also kind of rule based stemmer. It works by truncating all possible suffixes form and produce finite verb. Light stemming is used to find the representative indexing form of given word by the application of truncation of suffixes [10]. The core objective of light stemmer is to preserve the word meaning intact and so increases the retrieval performance of an IR system. A Light Stemmer Algorithm for Tamil Language is projected in Figure 3

| Light stemmer Algorithm for Tamil Language<br>Input : List of Tamil words<br>Output : Stemmed(Root) words |
|---|
| Step 1 : Eliminate the entire complex plural.<br>Step 2 : After the plural word is converted into singular word, during the iteration, the word is also checked for adjective; if it is found, then its equivalent verb is substituted. Example, the term ‘diya(ÊÂ)’ in Odiya(µÊÂ) will be changed to ‘du(Î)’ and the word is changed to ‘Oodu(µÎ)’.<br>Step 3 : After the adjectives are converted to main word, the tenses are eliminated such that Paadiya(À¡ÊÂ), Paadukinra(À¡Î¸¢ýÈ) and Paadum(À¡Îõ) will be changed to Paadu(À¡Î).<br>Step 4 : According to the identified suffix, the next possible suffix list is generated.<br>Step 5 : The Light algorithms are used for plural to singular conversion, and for adjective and tense words to main verbs conversion. |

Figure 3. A Light Stemmer Algorithm for Tamil Language

## V. EXPERIMENTAL ANALYSIS

The goal of experimental analysis is to calculate the accuracy of the proposed stemmer system can achieve. Parameters that can be used for evaluating Tamil stemmer algorithm in this proposed model is stemmer’s comparison performance and correctness of the stems produced by it. The detailed description of dataset used for experiment and analysis of experiment results and their discussions are under here.

### *A. Dataset*

Two test dataset are considered for evaluating the proposed algorithm. The test dataset I has 700 words which are collected from Tamil corpus (Central Institute of Indian Language). Test dataset 2 contains 1600 words constructed from the internet. The stem for these words have been defined manually. The training dataset consists of 3000 words taken from Tamil corpus. Table IV summarizes the details of training and test datasets.

TABLE IV DATA SET GENERATION

| Data set | Total Number of Words | Total Number of Unique Words | Minimum Length of the Word | Maximum Length of the Word |
|---|---|---|---|---|
| Training Dataset | 3000 | 847 | 3 | 14 |
| Test Dataset I | 700 | 234 | 3 | 9 |
| Test Dataset II | 1600 | 580 | 3 | 11 |

### *B. The Experiment Results and Discussion*

The experiment analyzed over two datasets in order to evaluate the performance of proposed algorithm. The accuracy is a parameter; it is used to evaluate the efficiency of the proposed Tamil stemmer algorithm over a rule based suffix stripping stemmer. The accuracy is defined in this proposed stemmer based on number of words stemmed correctly. The Table V shows the computational results of a rule based suffix stripping and light stemmer based on suffix rule generation. Table VI average result is generated from the Table V. From the Table VI, the maximum accuracy of rule based suffix stripping stemmer is 84.32% where as 86.73% maximum accuracy produced by light stemmer for the test dataset I. light stemmer achieves accuracy 2.41% more than the rule based suffix stripping stemmer. The accuracy of rule based suffix stripping stemmer in test dataset II is 78.4% and light stemmer accuracy is 79.83%. In the dataset II 1.43% more accuracy achieved by the light stemmer it is more than the rule based suffix stripping stemmer accuracy.

Accuracy is calculated from the equation (1). It is based on the Number of Words Stemmed correctly given by the Tamil Stemmer algorithms and number of unique words in the given datasets.

$$\text{Accuracy} = \text{Number of Correctly Stemmed words} / \text{Number of Unique words} * 100 \quad (1)$$

TABLE V TEST DATA

| Dataset | Number of Words | Number of Unique Words | Number of Correctly Stemmed Word | Accuracy | |
|---|---|---|---|---|---|
| | | | | Rule Based suffix Stripping algorithm | Light Stemmer algorithm |
| Dataset I | 200 | 37 | 30 | 81.0% | 85.9% |
| | 400 | 118 | 101 | 85.5% | 84.2% |
| | 600 | 182 | 152 | 83.5% | 87.6% |
| | 700 | 237 | 200 | 84.3% | 89.7% |
| Dataset II | 200 | 55 | 43 | 79.4% | 80.2% |
| | 400 | 148 | 116 | 78.8% | 78.4% |
| | 600 | 193 | 142 | 75.2% | 77.1% |
| | 800 | 391 | 290 | 74.1% | 81.2% |
| | 1000 | 424 | 338 | 79.7% | 79.3% |
| | 1200 | 486 | 400 | 82.2% | 82.1% |
| | 1400 | 539 | 402 | 74.6% | 76.2% |
| | 1600 | 580 | 480 | 82.8% | 84.0% |

TABLE VI AVERAGE RESULTS

| Data Set | Accuracy | |
|---|---|---|
| | A Rule Based Suffix Stripping Stemmer | A light Stemmer |
| Data Set I | 84.32% | 86.73% |
| Data Set II | 78.4% | 79.83% |
| Average | 81.36% | 83.28% |

The experiment is conducted over the dataset result of different accuracy (i.e. A1,A2,…) achieved by both rule based suffix stripping and light stemmer algorithm with required statistical parameters and their average result is shown in Table V. The performance analysis chart is presented among various parameters in Figure 4. The computational results clearly explain that the light stemmer algorithm is performing better than the rule based suffix stripping algorithm.

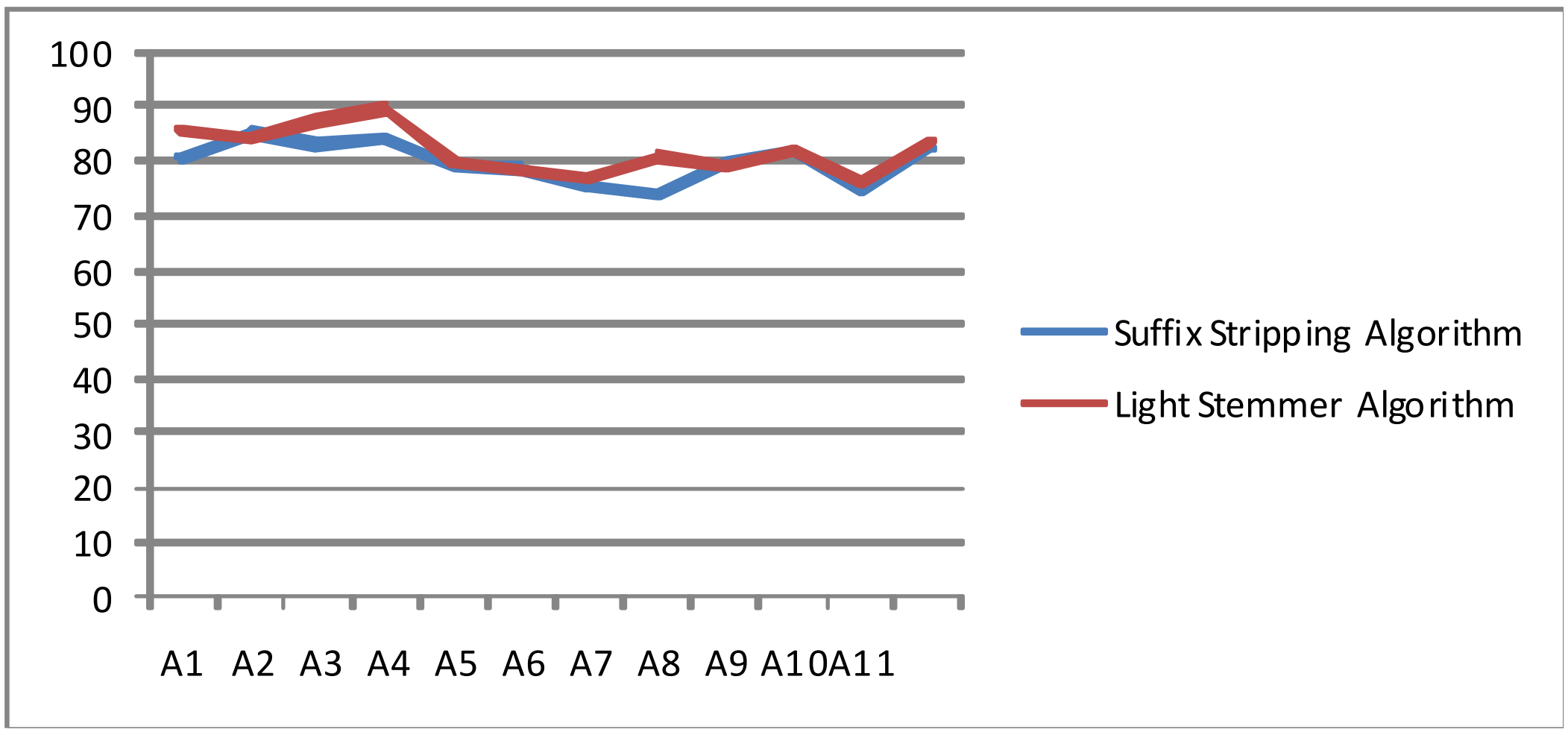

Figure 4. Performance Analyses for Tamil Stemmer Algorithms

## VI. CONCLUSION

Morphologically Tamil is a complex language since it has a number of variants in for a single word. Tamil language is rich in both inflectional and derivational morphologies. In this paper, a light stemmer for Tamil text is proposed to handle inflectional morphology word. This stemmer removes suffixes from a word to get stem word. From the computational result it is prove light stemmer approach is more suitable stemmer for Tamil language compare to a rule based suffix stripping stemmer.

### AUTHOR'S PROFILE

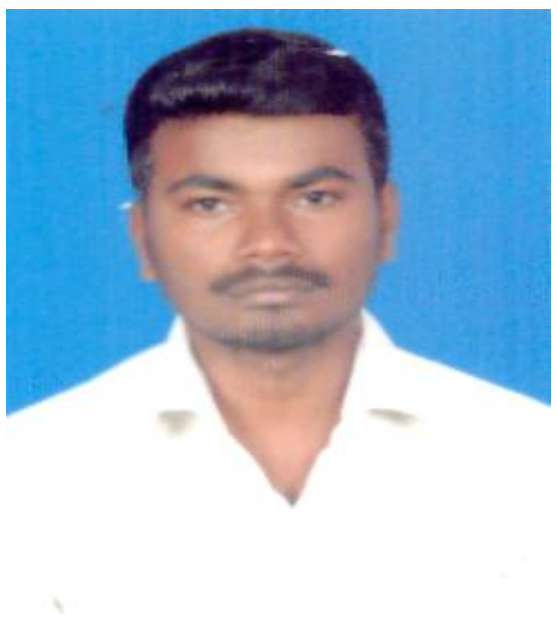

M.Thangarasu received Master of Computer Application degree from Bharathiar University. He purses M.Phil under supervision of Dr.R.Manavalan. His area of interest is data mining and Natural Processing Language (NLP).

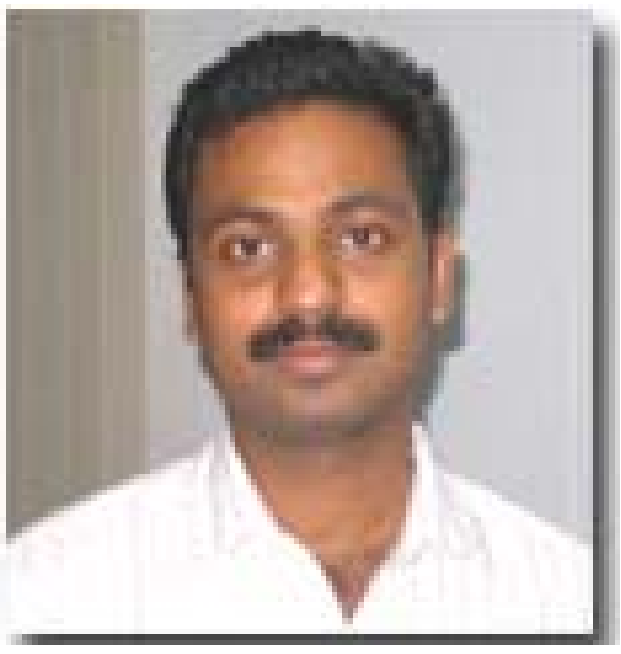

Dr. R. Manavalan is working as Assistant professor and Head in Department of computer science and Applications. He obtained Ph.D in Computer Science from Periyar University and published numerous research papers in international journals and also presented papers in various national and international conferences. His area of interest is soft computing, image processing and analysis, Theory of computation.